\documentclass[sigconf]{aamas18}  % do not change this line!

\usepackage{booktabs}

\setcopyright{ifaamas}  % do not change this line!
\acmDOI{doi}  % do not change this line!
\acmISBN{}  % do not change this line!
\acmConference[AAMAS'18]{Proc.\@ of the 17th International Conference on Autonomous Agents and Multiagent Systems (AAMAS 2018), M.~Dastani, G.~Sukthankar, E.~Andre, S.~Koenig (eds.)}{July 2018}{Stockholm, Sweden}  % do not change this line!
\acmYear{2018}  % do not change this line!
\copyrightyear{2018}  % do not change this line!
\acmPrice{}  % do not change this line!

\usepackage{comment}

\usepackage{booktabs,makecell}\usepackage{graphicx}
\usepackage{caption}
\usepackage{subcaption}
\usepackage{wrapfig}
\usepackage{tikz}
\usepackage{tabularx}
\usepackage{placeins}
\usepackage{dblfloatfix}
\usepackage{algorithm}
\usepackage{algpseudocode}
\usepackage{pifont}
\usepackage{bbding}
\usepackage{amsmath}
\usepackage{amssymb}
\usepackage{amsfonts}
\usepackage{amsthm}
\usepackage{mathtools}
\usepackage{bm}
\usepackage{mathrsfs}

\usepackage[hang,flushmargin]{footmisc}   % flush footnotes

\newcommand{\citeauthoryr}[1]{\citeauthor{#1}~\shortcite{#1}}

\algrenewcommand\algorithmicforall{\textbf{foreach}}
\algrenewcommand\algorithmicindent{.8em}

\newcommand{\Task}[1]{\mathcal{Z}^{(#1)}}

\newcommand{\st}[1]{\bm{s}^{(#1)}}

\newcommand{\Th}[1]{\bm{\theta}^{(#1)}}

\newcommand{\Xt}[1]{X^{(#1)}}

\newcommand{\yt}[1]{\bm{y}^{(#1)}}

\usepackage{xspace}

\newcommand{\Reals}{\ensuremath{\mathbb{R}}}

\usepackage{xspace}

\begin{document}

\title{Multi-Agent Distributed Lifelong Learning for\\ Collective Knowledge Acquisition}  % put your title here!
%\titlenote{Produces the permission block, and copyright information}

% AAMAS: as appropriate, uncomment one subtitle line; check the CFP
%\subtitle{Extended Abstract}
%\subtitle{Industrial Applications Track}
%\subtitle{Socially Interactive Agents Track}
%\subtitle{Blue Sky Ideas Track}
%\subtitle{Robotics Track}
%\subtitle{JAAMAS Track}
%\subtitle{Doctoral Mentoring Program}

%\subtitlenote{The full version of the author's guide is available as \texttt{acmart.pdf} document}

% AAMAS: submissions are anonymous for most tracks
%\author{Paper \#632}  % put your paper number here!

%% example of author block for camera ready version of accepted papers: don't use for anonymous submissions
%
\author{Mohammad Rostami}
\affiliation{
  \institution{University of Pennsylvania}
  \city{Philadelphia} 
  \state{PA}
  \country{USA} 
}
\email{mrostami@seas.upenn.edu}

\author{Soheil Kolouri}
\affiliation{%
 \institution{HRL Laboratories, LLC}
  \city{Malibu}
  \state{CA} 
  \country{USA} 
}
\email{skolouri@hrl.com}

\author{Kyungnam Kim}

\affiliation{%
  \institution{HRL Laboratories, LLC}
  \city{Malibu} 
  \state{CA}
  \country{USA}}
\email{kkim@hrl.com}

\author{Eric Eaton}
\affiliation{%
  \institution{University of Pennsylvania}
  \city{Philadelphia}
  \state{PA}
  \country{USA}}
  \email{eeaton@cis.upenn.edu}

%\author{Aparna Patel} 

%\affiliation{%
 %\institution{Rajiv Gandhi University}
 %\streetaddress{Rono-Hills}
% \city{Doimukh} 
% \state{Arunachal Pradesh}
% \country{India}}
%\author{Huifen Chan}
%\affiliation{%
%  \institution{Tsinghua University}
%  \streetaddress{30 Shuangqing Rd}
%  \city{Haidian Qu} 
%  \state{Beijing Shi}
%  \country{China}
%}

%\author{Charles Palmer}
%\affiliation{%
%  \institution{Palmer Research Laboratories}
%  \streetaddress{8600 Datapoint Drive}
%  \city{San Antonio}
%  \state{Texas} 
%  \postcode{78229}}
%\email{cpalmer@prl.com}
%
%\author{John Smith}
%\affiliation{\institution{The Th{\o}rv{\"a}ld Group}}
%\email{jsmith@affiliation.org}
%
%\author{Julius P.~Kumquat}
%\affiliation{\institution{The Kumquat Consortium}}
%\email{jpkumquat@consortium.net}
%
%% The example's default list of authors is too long for headers
%\renewcommand{\shortauthors}{B. Trovato et al.}

\begin{abstract}  % put your abstract here!
Lifelong machine learning methods acquire knowledge over a series of consecutive tasks, continually building upon their experience.  Current lifelong learning algorithms rely upon a single learning agent that has centralized access to all data. In this paper, we extend the idea of lifelong learning from a single agent to a network of multiple agents that collectively learn a series of tasks. Each agent faces some (potentially unique) set of tasks; the key idea is that knowledge learned from these tasks may benefit other agents trying to learn different (but related) tasks.  Our Collective Lifelong Learning Algorithm (CoLLA) provides an efficient way for a network of agents to share their learned knowledge in a distributed and decentralized manner, while preserving the privacy of the locally observed data. Note that a decentralized scheme is a subclass of distributed algorithms where a central server does not exist and in addition to data, computations are also distributed among the agents. We provide theoretical guarantees for robust performance  of the algorithm and empirically demonstrate that CoLLA outperforms existing approaches for distributed multi-task learning on a variety of data sets.
\end{abstract}

% AAMAS: the ACM CCS are not needed within AAMAS papers
%%
%% The code below should be generated by the tool at
%% http://dl.acm.org/ccs.cfm
%% Please copy and paste the code instead of the example below. 
%%
%\begin{CCSXML}
%<ccs2012>
% <concept>
%  <concept_id>10010520.10010553.10010562</concept_id>
%  <concept_desc>Computer systems organization~Embedded systems</concept_desc>
%  <concept_significance>500</concept_significance>
% </concept>
% <concept>
%  <concept_id>10010520.10010575.10010755</concept_id>
%  <concept_desc>Computer systems organization~Redundancy</concept_desc>
%  <concept_significance>300</concept_significance>
% </concept>
% <concept>
%  <concept_id>10010520.10010553.10010554</concept_id>
%  <concept_desc>Computer systems organization~Robotics</concept_desc>
%  <concept_significance>100</concept_significance>
% </concept>
% <concept>
%  <concept_id>10003033.10003083.10003095</concept_id>
%  <concept_desc>Networks~Network reliability</concept_desc>
%  <concept_significance>100</concept_significance>
% </concept>
%</ccs2012>  
%\end{CCSXML}
%
%\ccsdesc[500]{Computer systems organization~Embedded systems}
%\ccsdesc[300]{Computer systems organization~Redundancy}
%\ccsdesc{Computer systems organization~Robotics}
%\ccsdesc[100]{Networks~Network reliability}

  % put your semicolon-separated keywords here!

\maketitle

%%%%%%%%%%%%%%%%%%%%%%%%%%%%%%%%%%%%%%%%%%%%%%%%%%%%%%%%%%%%%%%%%%%%%%%%%%%%%%%%%%%%%%%%%%%%%%%%%%%%%%%%%
%% start of main body of paper

\section{Introduction}

Collective knowledge acquisition is common throughout different societies, from the collaborative advancement of human knowledge to the emergent behavior of ant colonies \cite{kao2014collective}.  It is the product of individual agents, each with their own interests and constraints, sharing and accumulating learned knowledge over time  in uncertain and dangerous real-world environments. Our work explores this scenario within machine learning and in particular, considers learning in a network of lifelong machine learning agents \cite{Thrun1996}.

Recent work in lifelong machine learning \cite{Ruvolo2013} has explored the notion of a single agent accumulating knowledge over its lifetime.  Such an individual lifelong learning agent reuses knowledge from previous tasks to improve its learning on new tasks, accumulating an internal repository of knowledge over time.  This lifelong learning process improves performance over all tasks, and permits the design of adaptive agents that are capable of learning in dynamic  environments.  Although current work in lifelong learning focuses on a single learning agent that incrementally perceives all task data, many real-world applications involve scenarios in which multiple agents must collectively learn a series of tasks that are distributed amongst them. Consider the following cases: 
 
\begin{itemize}
\item Multi-modal task data could only be partially accessible by each learning agent.  For example, financial decision support agents may have access only to a single data view of tasks or a portion of the non-stationary data distribution~\cite{he2011graph}.
\item Local data processing can be inevitable in some applications, such as when health care regulations prevent personal medical data from being shared between learning systems~\cite{zhang2012multi}.
\item Data communication may be costly or time consuming.  For instance, home service robots must process perceptions locally due to the volume of perceptual data, or wearable devices may have limited communication bandwidth~\cite{jin2015collaborating}.
\item As a result of data size or geographical distribution of data centers, parallel processing can be essential.  Modern big data systems often necessitates parallel processing in the cloud across multiple virtual agents, i.e. CPUs or GPUs~\cite{zhang2014evolutionary}.  
\end{itemize}

Inspired by the above scenarios, this paper explores the idea of {\em multi-agent lifelong learning}.  We consider multiple collaborating lifelong learning agents, each facing their own series of tasks, that transfer knowledge to collectively improve task performance and increase learning speed.  Note that this paper  does not address the privacy considerations that may arise from transferring knowledge between agents. Also, despite potential extendability to parallel processing systems, our focus here is more on  collaborative agents that receive sequential tasks. Existing methods in the literature have mostly investigated special cases of this setting for distributed multi-task learning (MTL)~\cite{chen2014multitask,parameswaran2010large,jin2015collaborating}.

To develop multi-agent distributed lifelong learning, we follow a parametric approach and formulate the learning problem as an online MTL optimization over a network of agents. Each agent seeks to learn parametric models for its own series of (potentially unique) tasks.  The network topology   imposes communication constraints among the agents. For each agent, the corresponding task model parameters are represented as a task-specific sparse combination of atoms of its local knowledge base \cite{kumar2012learning,Ruvolo2013,Maurer2013}. The local knowledge bases allow for knowledge transfer from learned tasks to the future tasks for each individual agent.  The agents share their knowledge bases with their neighbors, update them to incorporate the learned knowledge representations of their neighboring agents, and come to a local consensus to improve learning quality and speed. Our contribution is to use the Alternating Direction Method of Multipliers (ADMM) algorithm~\cite{boyd2011distributed} to solve this global optimization problem in an online distributed setting; our approach decouples this problem into  local optimization problems that are individually solved by the agents.  ADMM allows for transferring the learned local knowledge bases without sharing the specific learned model parameters among neighboring agents. We propose an algorithm with nested loops to allow for keeping the procedure both online and distributed. We call our approach the Collective Lifelong Learning Algorithm (CoLLA).  We provide theoretical analysis on the convergence of CoLLA and   empirically validate the practicality of the proposed algorithm on variety of standard MTL benchmark datasets.

 \section{Related Work}
This paper considers scenarios where multiple lifelong learning agents learn a series of tasks distributed among them. Each agent shares high level information with its neighboring agents, while processing data privately. Our approach draws upon various subfields of machine learning, which we briefly survey below.
 
\noindent\textbf{Multi-Task and Lifelong Learning:~~}
Multi-task learning (MTL) \cite{caruana1997} seeks to share knowledge among multiple related tasks.  Compared to single task learning (STL), MTL increases generalization performance and reduces the data requirements for learning tasks through benefiting from task similarities.  One major challenge in MTL is modeling task similarities to selectively share information and enable knowledge transfer between tasks \cite{caruana1997}. If this process identifies incorrect task relationships, sharing knowledge can degrade performance through the phenomenon of negative transfer. Various techniques have been developed to model task relations, including modeling a task distance metric \cite{ben2007analysis}, using correlations to determine when transfer is appropriate \cite{wang2014adaptive}, and regularizing task parameters \cite{Argyriou}. An effective parametric approach is to group similar tasks by assuming that task parameters can be represented sparsely in a dictionary domain. Then, by imposing sparsity on task-specific parameters, similar tasks can be grouped together enabling positive knowledge transfer and the learned dictionary would model task relations \cite{kumar2012learning}. Upon learning the dictionary, similar tasks would share a subset of dictionary columns which helps to avoid negative transfer.
 
Lifelong learning   is a special case of MTL in which an agent receives tasks consecutively and continuously and learns the tasks in an online sequential setting.  To improve learning  performance on each new task, the agent transfers  knowledge obtained from the previous tasks, and then stores new or revised knowledge for future use~\cite{Pan2010a}. As a result, the learning agent is able to adapt itself to dynamic environments and drifts in data distribution, and consistently improves its performance. This ability is essential for emerging applications such as personal intelligent robot assistants and chatbots which continuously interact with many different users. Pioneer works in lifelong learning are built upon extending MTL schemes to an online setting. \citeauthoryr{Ruvolo2013} extended the MTL method proposed by \citeauthoryr{kumar2012learning} to a lifelong learning setting, creating an efficient and fast algorithm for lifelong learning. Our approach is partially based upon their formulation, which serves as the foundation to develop our novel collective lifelong learning framework.  Note that unlike our work, most prior MTL and lifelong learning work consider  the case where all tasks are accessible by a single agent in a centralized scheme.  

\noindent\textbf{Distributed Machine Learning:~~} There has been a growing interest in developing scalable learning algorithms using distributed optimization \cite{zhang2014asynchronous}, motivated by the emergence of big data \cite{cevher2014convex}, security and privacy  constraints \cite{yan2013distributed}, and the notion of cooperative and collaborative learning agents \cite{chen2012diffusion}.  Distributed machine learning allows multiple agents to collaboratively mine information from large-scale data. The majority of these settings are graph-based, where each node in the graph represents a portion of data or an agent. Communication channels between the agents then can be modeled via edges in the graph. Some approaches   assume there is a central server (or a group of server nodes) in the network, and the worker agents transmit locally learned information to the server(s), which then perform knowledge fusion \cite{xing2015petuum}. Other approaches assume that processing power is distributed among the agents, which exchange information with their neighbors during the learning process \cite{chen2014multitask}.  We formulate our problem in the latter setting, as it is less restrictive. Following the dominant paradigm of distributed optimization, we also assume that the agents are synchronous.

These algorithms formulate learning as an optimization problem over the network and use techniques from distributed optimization to acquire the global solution. Various techniques have been explored, including stochastic gradient descent \cite{xing2015petuum}, proximal gradients  \cite{li2014communication}, and ADMM \cite{xing2015petuum}. Within the ADMM framework, it is assumed that the objective function over the network can be decoupled into a sum of independent local functions for each node (usually risk functions) \cite{mairal2009online}, constrained by network topology. Through a number of iterations on primal and dual variables of the Lagrangian function, each node solves a local optimization, and then through information exchange, constraints imposed by the network are realized by updating the dual variable.   In scenarios where maximizing a cost for some agents translate to minimizing the cost for others (e.g., adversarial games), game theoretical notions are used to define a global optimal state for the agents \cite{li2013designing}.

\noindent\textbf{Distributed Multi-task Learning:~~}
Although it seems natural to consider MTL agents that collaborate on   related tasks, most prior distributed learning work focuses on the setting where all agents try to learn a single task. Only recently have MTL scenarios been investigated where the tasks are distributed \cite{jin2015collaborating,mateos2015distributed,wang2015distributed,baytas2016asynchronous,xie2017privacy,liu2017distributed}. In such a  setting,  data  must  not be  transferred  to  a  central  node because of communication and privacy/security constraints.  Only  the learned models or high-level information can be exchanged by neighboring agents. Distributed MTL has also been explored in reinforcement learning settings \cite{el2017scalable} where the focus is on developing a scalable multitask
policy search algorithm.  These distributed MTL methods are mostly  limited to off-line (batch) settings where each agent handles only one task \cite{mateos2015distributed,wang2015distributed}. \citeauthoryr{jin2015collaborating} consider an online setting, but require the existence of a central node, which is restrictive. In contrast, our work considers decentralized and distributed multi-agent MTL in a lifelong learning setting, without the need for a central server.  Moreover, our approach employs homogeneous agents that collaborate to improve collective performance over consecutive distributed tasks in a lifelong learning setting. This can be considered as a special case of concurrent learning where learning task concurrently by multiple agents can speed up learning rate~\cite{jansen2003exploring}.
 
Similar to prior works \cite{mateos2015distributed,wang2015distributed,el2017scalable}, we use    distributed optimization  to tackle  the collective lifelong learning problem. These existing approaches can only handle an off-line setting where all the task data is available in a batch for each agent. In contrast, we propose an online learning procedure which can address consecutively arriving tasks.  In each iteration, the agents receive and learn their local task models. Since the agents are synchronous, once the tasks are learned by agents, a message-passing scheme is then used to transfer and update knowledge between the neighboring agents in each iteration. In this manner, knowledge will disseminate among all agents over time, improving collective performance. Similar to most distributed learning settings, we assume there is a latent knowledge base that underlies all tasks and that each agent is trying to learn a local version of that knowledge base based on its own (local) observations and  knowledge exchange with neighboring agents, modeled by edges (links) of the representing network graph.

\section{Lifelong Machine Learning}
We consider a set of $T$ related (but different) supervised regression or classification tasks, each with labeled training data, i.e. $\left\{\Task{t} = \left( \Xt{t}, \yt{t} \right)\right\}_{t=1}^T$,  where $\Xt{t}=[\bm{x}_1,\ldots,\bm{x}_M]\in \mathbb{R}^{d\times M}$ represents $M$ task data instances characterized by $d$ features, and $\yt{t}=[y_1,\ldots,y_m]^\top\in \mathcal{Y}^M$ are the corresponding targets. Typically, $\mathcal{Y}=\{\pm 1\}$ for binary classification tasks and $\mathcal{Y}= \mathbb{R}$ for regression tasks. We assume that for each task, the hidden mapping from each data point $\bm{x}_m$ to the corresponding target $y_m$, $f:\mathbb{R}^{d}\rightarrow \mathcal{Y}$ can be parametrized as $y_m= f(\bm{x}_m;\Th{t})$, where $\Th{t}\in \mathbb{R}^d$. In this work, we consider a linear mapping $ f(\bm{x}_m;\Th{t})=\langle\Th{t}\cdot \bm{x}_m\rangle$ where $\Th{t}\in \mathbb{R}^d$, but our framework is readily generalizable to nonlinear parametric mappings (e.g., using generalized dictionaries \cite{wang2016generalized}).  After receiving a task $\Task{t}$, the goal of the agent is to learn the mapping $f(\bm{x}_m;\Th{t}) $ by estimating the corresponding optimal task parameter $\Th{t}$ using the   training data such that it well-generalizes on testing data points from that task.
An agent can learn the task models by solving for the  optimal parameters  \mbox{$\bm{\Theta^*}=[\Th{1},\ldots,\Th{T}]$} in the following expected risk minimization (ERM) problem:
\begin{equation} \label{eqn:LMLObjective} 
	\min_{\bm{\Theta}} \frac{1}{T} \sum_{t=1}^T \mathbb{E}_{\Xt{t}\sim\mathcal{D}^{(t)}}\left(\mathcal{L}\left(\Xt{t}, \yt{t};\Th{t}\right)\right)  + \Omega(\bm{\Theta})\enspace,
\end{equation}
where $\mathcal{L}(\cdot)$ is a   loss function for measuring data fidelity,   $\mathbb{E}(\cdot)$ denotes the expectation on the task's data distribution $\mathcal{D}^{(t)}$, and $\Omega(\cdot)$ is a regularization function that models task relations by coupling model parameters to transfer knowledge among the tasks. Almost all parametric MTL, online, and lifelong learning algorithms solve instances of \eqref{eqn:LMLObjective} given a particular  form of  $\Omega(\cdot)$ to impose a specific coupling scheme and an  optimization mode, i.e. online or   offline.

To model task relations, the GO-MTL algorithm \cite{kumar2012learning} uses classic Empirical Risk Minimization (ERM) to estimate the expected loss and solves the objective \eqref{eqn:LMLObjective} by assuming that the task parameters  can be decomposed into a shared knowledge dictionary base  $\bm{L}\in\mathbb{R}^{d\times u}$ to facilitate knowledge transfer and task-specific sparse coefficients $\st{t}\in \mathbb{R}^u$, such that  $\Th{t} = \bm{L}\st{t}$. In this factorization, the hidden structure of the tasks is represented in the knowledge dictionary base and similar tasks are grouped by imposing sparsity on $\st{t}$'s. Tasks that use the same columns of the dictionary are clustered to be similar, while tasks that do not share any column can be considered belonging to different groups. In other words, more overlap on sparsity patterns of two tasks imply more similarity between those two tasks.This factorization  has been analytically shown  to enable knowledge transfer when dealing with related tasks by grouping the similar tasks~\cite{kumar2012learning,Maurer2013}. Following this assumption and employing ERM, the  objective \eqref{eqn:LMLObjective}  can be estimated as: 
\begin{equation} \label{eqn:MTLObjective}
	\!\min_{\bm{L},\bm{S}} \frac{1}{T} \!\sum_{t=1}^T\!\left[\!\hat{\mathcal{L}}\left(\Xt{t}\!, \yt{t}\!,\bm{L}\st{t}\right) + \mu \| \st{t} \|_1 \!\right] + \lambda\|\bm{L}\|_\textsf{F}^2\enspace,
\end{equation}
where $\bm{S} = [\st{1} \cdots\ \st{T}]$ is the matrix of sparse vectors, $\hat{\mathcal{L}}(\cdot)$  is the empirical loss function on task training data,  $\|\cdot\|_{\mathsf{F}}$ is the Frobenius norm to regularize complexity and impose uniqueness, $\|\cdot\|_1$ denotes the $L_1$ norm to impose sparsity on $\st{t}$, and $\mu$ and $\lambda$ are regularization parameters. Eq. \eqref{eqn:MTLObjective} is not a convex problem in its general form, but for a convex loss function  is convex in each individual optimization variables $\bm{L}$ and $\bm{S}$. Given all tasks' data in a batch, Eq. \eqref{eqn:MTLObjective} can be solved in an offline batch-mode by an iterative alternating optimization scheme~\cite{kumar2012learning}. In each alternation step, Eq. \eqref{eqn:MTLObjective} is solved to update a single variable by treating the other variable to be constant. This scheme leads to an MTL algorithm that enables the agent to share information selectively among the tasks.

Solving Eq. \eqref{eqn:MTLObjective} in an offline setting is not applicable for lifelong learning because all tasks are not available in a single batch.  A  lifelong learning agent   faces   tasks sequentially over its lifetime, where each task  should be learned efficiently and fast using knowledge transfered from past experience. The agent does not   know the total number of
tasks, nor the order the tasks are received. In other words, for each task $\Task{t}$, the corresponding parameter $\Th{t}$ is learned using knowledge obtained from the  training
data from previous tasks $\{\Task{1}, \dots, \Task{t-1}\}$ and upon learning $\Task{t}$, the learned or updated knowledge is stored to benefit  learning future tasks in an online scheme.   To solve  Eq.~\eqref{eqn:MTLObjective} in an online lifelong learning setting, \citeauthoryr{Ruvolo2013} first approximate the loss function $\mathcal{L}(\Xt{t}, \yt{t},\bm{L}\st{t})$ using a second-order Taylor expansion of the loss function around a single-task ridge-optimal parameters.  This technique reduces the  objective function of Eq.~\eqref{eqn:MTLObjective} to the problem of online dictionary learning \cite{mairal2009online} as follows:
 \begin{align}  \label{eqn:SparseObjective}
&	\min_{\bm{L}} \frac{1}{T} \sum_{t=1}^TF^{(t)}(\bm{L}) + \lambda\|\bm{L}\|_\textsf{F}^2\enspace,\\&  \label{eqn:localbjective}
 F^{(t)}(\bm{L})=\min_{\st{t} }\left[ \| \bm{\alpha}^{(t)} -\bm{L}\st{t}\|^2_{\bm{\Gamma}^{(t)}} +  \mu \| \st{t} \|_1 \right]\enspace, 
 \end{align} 
 where $\|\bm{x}\|^2_{\bm{A}} = \bm{x}^{\!\top}\!\!\bm{A}\bm{x}$,    %$\bm{\alpha}^{(t)}=\arg\min_{\Xt{t}, \yt{t}\Th{t}}\{\hat{\mathcal{L}}(\Xt{t}, \yt{t},\Th{t}). 
 %+\gamma\|\Th{t}\|_2^2 \}\enspace$, 
     $\bm{\alpha}^{(t)} \in \Reals^d$ is the ridge estimator for task $\Task{t}$:
 \begin{equation}
 \bm{\alpha}^{(t)}=\arg\min_{\Th{t}}\left[\hat{\mathcal{L}}(\Th{t}) 
 +\gamma\|\Th{t}\|_2^2 \right]\enspace,
 \end{equation}
 and $ \bm{\Gamma}^{(t)}$ is the Hessian of the loss $\hat{\mathcal{L}}(\cdot)$ at $\bm{\alpha}^{(t)}$ which is assumed to be strictly positive definite. Also, $\gamma\in\mathbb{R}^d$ is the ridge regularization parameter. To solve  Eq.~\eqref{eqn:SparseObjective} in an online setting, still an alternation scheme is used but
when a new task arrives, only the corresponding sparse vector $\st{t}$ for that tasks is computed using $\bm{L}$  to update the sum $ \sum_{t=1}^T F(\bm{L})$. In this setting, Eq. \eqref{eqn:localbjective} becomes a task-specific online operation that leverages knowledge transfer.   Finally the shared basis $\bm{L}$ is updated via Eq. \eqref{eqn:SparseObjective} to store the learned knowledge from $\Task{t}$ for future use. Despite using Eq. \eqref{eqn:localbjective} as an approximation to solve for $\bm{s}$, \citeauthoryr{Ruvolo2013} proved that the learned knowledge base $\bm{L}$ stabilize as more tasks are learned   and would eventually converge to the offline scheme solution of~\citeauthoryr{kumar2012learning}. Moreover, the solution of Eq.~\eqref{eqn:LMLObjective}  converges almost surely to the solution of Eq.~\eqref{eqn:MTLObjective} as $T\rightarrow \infty$.
While this technique leads to an efficient and fast algorithm for lifelong learning, it requires centralized access to all tasks' data by a single agent. The approach we explore, COLLA, benefits from the idea of the second-order Taylor approximation and online optimization scheme proposed by \citeauthoryr{Ruvolo2013}, but eliminates the need for a centralized data access. CoLLA achieves a distributed and decentralized knowledge update by formulating a multi-agent lifelong learning optimization problem over a network of collaborating lifelong learning agents. The resulting optimization problem, can be solved in a distributed setting, enabling collective learning, as we describe next.

\section{Multi-Agent Lifelong Learning}
We consider a network of $N$   collaborating lifelong learning agents with an arbitrary order on  the agents. Note however this arbitrary order is restrictive and needs to be known and fixed. Each agent receives a potentially unique tasks at each time-step in a lifelong learning setting. There is also some true underlying hidden knowledge base  for all tasks, and each agent learns a local view of this knowledge base based on its own task distribution.  We assume that each agent $i$ solves a local version of the objective \eqref{eqn:SparseObjective} to estimate its own local knowledge base $\bm{L}_i$. We also assume that the agents are synchronous, meaning that at each time step, they simultaneously receive and learn one task.   We model the communication mode of these agents by an undirected graph $\mathcal{G}=(\mathcal{V},\mathcal{E})$, where the set of static nodes $\mathcal{V}=\{1,\ldots, N\}$ denotes the agents and the set of edges $\mathcal{E} \subset \mathcal{V}\times \mathcal{V}$, with $|\mathcal{E}|=e$, specifies possibility of communication between the agents.  We assume for each edge $(i,j)\in \mathcal{E} $, the nodes $i$ and $j$ are connected or they can communicate information, with $j>i$ for uniqueness and set orderability. We define the neighborhood $\mathcal{N}(i)$ of node $i$ as the set of all nodes that are connected to it.  To allow for knowledge and information flow between all the agents, we further assume that the network graph is connected.  Note that all nodes are assumed to be homogeneous and as a result there is no central node that guides collaboration among the agents.

We use the graph structure to formulate a lifelong machine learning problem on this network.  Although each agent learns its own individual dictionary, we encourage local dictionaries of neighboring nodes (agents) to be similar by adding a set of soft equality constraints on neighboring dictionaries, i.e. $ \bm{L}_i = \bm{L}_j, \forall (i,j)\in \mathcal{E}$. We can represent all these constraints as a single linear combination on local dictionaries.  It is easy to show these $e$ equality constraints  can be written compactly as $(\bm{H} \otimes \bm{I}_{d\times d})\tilde{\bm{L}}= \bm{0}_{ed\times u}$, where $\bm{H}\in\mathbb{R}^{e\times N}$ is the node arc-incident matrix of $\mathcal{G}$ (for a given row $1\le l\le e$, corresponding to    the  $l^{th}$ edge $(i,j)$, $H_{lq}=0$ except for $H_{li}=1$ and $H_{lj}=-1$), $\bm{I}_{d\times d}$ is the identity matrix, $\tilde{\bm{L}}=[\bm{L}^\top_1,\ldots,\bm{L}^\top_N]^\top$, and $\otimes$ denotes   the  Kronecker product. 
Let $\bm{E}_i\in\mathbb{R}^{ed\times d}$ be a column partition of $\bm{E}=(\bm{H}\otimes \bm{I}_d)=[\bm{E}_1,...,\bm{E}_N]$; we can  compactly write the $e$ equality constraints as $\sum_i \bm{E}_i\bm{L}_i=\bm{0}_{ed\times u}$.
Each of $\bm{E}_i\in\mathbb{R}^{de\times d}$ matrices is a tall block matrix consisting of $d\times d$ blocks, $\{[\bm{E}_i]j\}_{j=1}^e$, that are either zero matrix $\forall j\notin \mathcal{N}(i)$, $\bm{I}_d, \forall j\in \mathcal{N}(i), j>i$, or $-\bm{I}_d, \forall j\in \mathcal{N}(i), j<i$. Note that $\bm{E}_i^\top\bm{E}_j=\bm{O}_d$ if $j \notin \mathcal{N}(i)$, where is $ d\times d $  zero matrix.
Following this notation, we can  reformulate    the MTL objective  \eqref{eqn:SparseObjective} for multiple agents as the following linearly constrained  optimization problem over the network graph $\mathcal{G}$:
 \begin{eqnarray} \label{eqn:ADMMSparseObjective}
 &\min_{\bm{L}_1,\ldots\bm{L}_N}& \frac{1}{T} \sum_{t=1}^T   \sum_{i=1}^N F_i^{(t)}(\bm{L_i}) + \lambda\|\bm{L_i}\|_\textsf{F}^2\nonumber \\&\text{s.t.}& \sum_{i=1}^N \bm{E}_i\bm{L}_i=\bm{0}_{ed\times u}\enspace.
\end{eqnarray} 
 Note that in Eq.~\eqref{eqn:ADMMSparseObjective} optimization variables are not coupled by a global variable and hence in addition to being a distributed problem, Eq.~\eqref{eqn:ADMMSparseObjective} is also a decentralized problem. In order to deal with the dynamic nature and time-dependency of the objective \eqref{eqn:ADMMSparseObjective}, we assume that at each time step $t$, each agent receives a task and computes $F_i^{(t)}(\bm{L_i})$ locally via \eqref{eqn:localbjective}  based on this local task. Then, through $K$ information exchanges during that   time step,   the local
dictionaries are updated such that the agents reach a local consensus and hence benefit from all the tasks that are received by the network in that time step.
To split the constrained objective \eqref{eqn:ADMMSparseObjective} into a sequence of local unconstrained agent-level problems, we use   the extended ADMM algorithm \cite{mairal2009online,mota2013d}.   This algorithm generalizes ADMM~\cite{boyd2011distributed} to account for linearly constrained  convex  problems with   a sum of $N$ separable objective functions.  Similar to ADMM, we  first need to form the augmented Lagrangian $\mathcal{J}_T(\bm{L}_1,\ldots,\bm{L}_N, \bm{Z})$  for  problem     \eqref{eqn:ADMMSparseObjective} at time $t$ in order to replace the constrained problem by an unconstrained objective function which has an added penalty term:
 \begin{equation} \label{eqn:AugADMMObjective}
 \begin{split}
 \mathcal{J}_T&(\bm{L}_1,\ldots,\bm{L}_N, \bm{Z})= \frac{1}{T} \sum_{t=1}^T   \sum_{i=1}^N F_i^{(t)}(\bm{L_i}) +  \\&\lambda\|\bm{L_i}\|_\textsf{F}^2+  \langle\bm{Z},\sum_{i=1}^N\bm{E}_i\bm{L}_i\rangle+\frac{\rho}{2} \|\sum_{i=1}^N\bm{E}_i\bm{L}_i\|_\textsf{F}^2\enspace,
\end{split}
\end{equation} 
where $\langle\bm{Z},\sum_{i=1}^N\bm{E}_i\bm{L}_i\rangle=\text{tr}(\bm{Z}^{\top}\sum_{i=1}^N\bm{E}_i\bm{L}_i)$ denotes   the matrix trace  inner product, $\rho\in\mathbb{R}^+$ is    a regularization  penalty  term parameter for violation of the constraint. Finally, the block matrix $\bm{Z}=[\bm{Z}_1^\top,\ldots,\bm{Z}_e^\top]^\top\in \mathbb{R}^{ed\times u}$ is the ADMM dual variable. The extended ADMM algorithm solves Eq.\eqref{eqn:ADMMSparseObjective} by iteratively updating the dual and primal variables using the following local split iterations:
 \begin{align}  
  \bm{L}_1^{k+1}&=\text{argmin}_{\bm{L}_1}\mathcal{J}_T(\bm{L}_1,\bm{L}_2^k\ldots,\bm{L}_N^k, \bm{Z}^k)\enspace,\nonumber\\
 \bm{L}_2^{k+1}&=\text{argmin}_{\bm{L}_2}\mathcal{J}_T(\bm{L}_1^{k+1},\bm{L}_2,\ldots,\bm{L}_N^k, \bm{Z}^k),\enspace\nonumber\\&~~\vdots \label{eqn:ADMMstepsLa}\\
\bm{L}_N^{k+1}&=\text{argmin}_{\bm{L}_N}\mathcal{J}_T(\bm{L}_1^{k+1},\bm{L}_2^{k+1},\ldots,\bm{L}_N, \bm{Z}^k),\enspace\nonumber\\
 \bm{Z}^{k+1}&=\bm{Z}^k+\rho(\sum_{i=1}^N\bm{E}_i\bm{L}_i^{k+1}) \enspace. \label{eqn:ADMMstepsZ} 
\end{align} 
The first $N$ problems are primal agent-specific problems to update each local dictionary    and the last problem  updates the dual variable.
These iterations  split the objective \eqref{eqn:AugADMMObjective} into local primal optimization problems to update    each of the  $\bm{L}_i$'s, and then synchronize the agents to share information through updating the dual variable. 

Note that the $j$'th column of $\bm{E}_i$ is only nonzero when $j\in\mathcal{N}(i)$ $[\bm{E}_i]_j=\bm{0}_d, \forall j\notin \mathcal{N}(i)$, hence the update rule for the dual variable is indeed $e$ local block updates by   agents that share an edge:
\begin{equation}\label{eqn:dualupdate}
\bm{Z}^{k+1}_l=\bm{Z}^k_l+\rho(\bm{L}_i^{k+1} - \bm{L}_j^{k+1}) \enspace,
\end{equation}
for   the $l^{th}$ edge (i,j). This means that to update the dual variable, agent $i$ solely needs to keep track of  copies of those blocks $\bm{Z}_l$ that are shared with neighboring agents, reducing  \eqref{eqn:ADMMstepsZ} to a set of distributed local operations. Note that iterations in \eqref{eqn:ADMMstepsLa} and \eqref{eqn:dualupdate} are performed $K$ times at each instance $t$ for each agent to allow for agents  to converge to a stable solution.  At each time    step  $t$, the stable solution from the previous  time step  $t-1$ is used to initialize dictionaries and the dual variable in \eqref{eqn:ADMMstepsLa}. Due to convergence guarantees of extended ADMM~\cite{mairal2009online}, this simply means that at each iterations all the tasks that are received by the agents are considered to update the knowledge bases.

\subsection{ Dictionary Update Rule }

Splitting an optimization using ADMM is particularly helpful if optimization on primal variables can be solved efficiently, e.g. closed form solution. We show that the local primal updates in \eqref{eqn:ADMMstepsLa}   can be solved in closed form. We simply compute and then null the gradients of the primal problems, which leads to systems of linear problems for each local dictionary $\bm{L}_i$:
 \begin{equation} \label{eqn:ADMMObjectivegrad}
 \small
 \begin{split}
 \!\!\frac{\partial \mathcal{J}_T}{\partial \bm{L}_i} &=  
 \frac{1}{T}   \sum_t \frac{\partial F^{(t)}}{\partial \bm{L}_i}+ \\& \rho 
 \bm{E}_i^\top\big(\bm{E}_i\bm{L}_i+\sum_{j,j>i}\bm{E}_j\bm{L}_j^{k}+\sum_{j,j<i}\bm{E}_j\bm{L}_j^{k+1}+\frac{1}{\rho}\bm{Z}\big) + 2\lambda \bm{L_i} \\&=  
\frac{2}{T}\sum_{t=1}^T\bm{\Gamma}_i^{(t)} (\bm{L}_i\bm{s}_i^{(t)}- \bm{\alpha}_i^{(t)})\bm{s}_i^{(t)\top}+ \\ \!\!& 
\bm{E}_i^\top\!\big(\bm{E}_i\bm{L}_i+\!\!\sum_{j,j>i}\!\!\bm{E}_j\bm{L}_j^{k}+\!\!\sum_{j,j<i}\!\!\bm{E}_j\bm{L}_j^{k+1}\!+\frac{1}{\rho}\bm{Z}\big) + 2\lambda \bm{L_i} \ \! \\&=
 \frac{2}{T}\sum_t\bm{\Gamma}_i^{(t)} \bm{L}_i\bm{s}_i^{(t)}\bm{s}_i^{(t)\top}+ \\& (\rho |\mathcal{N}(i)|+2\lambda)\bm{L}_i+\sum_{j\in \mathcal{N}(i)}\bm{E}_i^\top\bm{Z}_j-\frac{2}{T}\sum_t\bm{\Gamma}_i^{(t)} \bm{\alpha}_i^{(t)} \bm{s}_i^{(t)\top}\\&\rho \big( \sum_{ j<i,j\in\mathcal{N}(i)}\bm{E}_i^\top\bm{E}_j\bm{L}_j^{k+1}+   \sum_{ j>i,j\in\mathcal{N}(i)}\bm{E}_i^\top\bm{E}_j\bm{L}_j^{k} \big)=0\!.
\end{split}
\end{equation} 
Note that despite our compact representation, primal iterations in \eqref{eqn:ADMMstepsLa}  involve only dictionaries  from neighboring agents ($\forall j\notin\mathcal{N}(i)$ because $\bm{E}_i\bm{E}_j=0$ and  $[\bm{E}_i]j=\bm{0}_d, \forall j\notin \mathcal{N}(i)$). Moreover, only  blocks of dual variable $\bm{Z}$ that correspond to  neighboring agents are needed to update each knowledge base. This means that iterations in \eqref{eqn:ADMMObjectivegrad} are also fully distributed and decentralized local operations. 

To solve for $\bm{L_i}$, we vectorize  both sides of Eq.~\eqref{eqn:ADMMObjectivegrad} and then after applying a property of Kronecker (($\bm{B}^\top \otimes \bm{A})\text{vec}(\bm{X})= \text{vec}(\bm{AXB})$) on line 5 of the equation, Eq. \eqref{eqn:ADMMObjectivegrad} simplifies to the following   linear equation update rules for local knowledge base dictionaries:  
 \begin{align} 
&\bm{A}_i = \big(\frac{\rho}{2}|\mathcal{N}(i)|+\lambda\big) \bm{I}_{dk}  +\frac{1}{T}\sum_{t=1}^T (\bm{s}_i^{(t)}\bm{s}_i^{(t)\top})\otimes\bm{\Gamma}_i^{(t)},\enspace    \nonumber\\
&\bm{b}_i= \text{vec} \biggl(\frac{1}{T}\sum_{t=1}^T\bm{s}_i^{(t)\top}\otimes  (\bm{\alpha}_i^{(t) \top} \bm{\Gamma}_i^{(t)}  )-\frac{1}{2}\sum_{j\in\mathcal{N} (i)} \bm{E}_i^{\top} \bm{Z}_j\nonumber\\&\hspace{3mm}-\frac{\rho}{2} \big( \sum_{ j<i,j\in\mathcal{N}(i)}\bm{E}_i^\top\bm{E}_j\bm{L}_j^{k+1}+   \sum_{ j>i,j\in\mathcal{N}(i)}\bm{E}_i^\top\bm{E}_j\bm{L}_j^{k} \big)\biggr),  \nonumber\\
\bm{L_i} &\leftarrow \text{mat}_{d,k} \bigl(  \bm{A}_i^{-1} \bm{b}_i \bigr)  \label{eqn:updateruleL} \enspace,
\end{align} 
where   $\text{vec}(\cdot)$ denotes  the matrix to vector (via column stacking)  and $\text{mat}(\cdot)$ denotes    the  vector to matrix operations. To avoid the sums on $t$ and storing learned tasks data, we construct both $\bm{A}_i$   and $\bm{b}_i$ incrementally as tasks are learned. Our    method,    the  Collective Lifelong Learning Algorithm (CoLLA), is summarized in Algorithm~\ref{CoLLA}.

\begin{algorithm}[tb!]
\caption{$\mathrm{CoLLA}\left (k,d,\lambda,\mu, \rho \right)$\label{CoLLA}} 
\begin{algorithmic}[1]
\State $T \gets 0, \enspace \bm{A}\gets \bm{zeros}_{k  d,k  d}$,
\State $\bm{b} \gets \textbf{zeros}_{k ,1}$, \enspace $\bm{L}_i \gets \textbf{zeros}_{d,k}$
\While{$\mbox{MoreTrainingDataAvailable()}$}
\State $T \gets T + 1$
\While{ $i\leq N$}
\State $\left (\mathbf{X}_i^{(t)},\mathbf{y}_i^{t},t \right) \gets \mbox{getTrainingData()}$
\State $\left (\bm{\alpha}^{(t)}_i, \bm{\Gamma}_i^{(t)} \right) \gets \mbox{singleTaskLearner}(X^t,y^t)$
\State  $\bm{s}^{(t)}_i \gets$ Equation~\ref{eqn:localbjective}
\While{$k\le K$} 
\State $\bm{A}_i \gets \bm{A}_i + (\bm{s}_i^{(t)}\bm{s}_i^{(t)\top})\otimes\bm{\Gamma}_i^{(t)} $
\State $\bm{b}_i \gets \bm{b}_i +\text{vec} \biggl(\bm{s}_i^{(t)\top}\otimes  (\bm{\alpha}_i^{(t) \top} \bm{\Gamma}_i^{(t)}  ) \biggr)$
\State $\bm{L}_i \gets \mbox{reinitializeAllZero}(\bm{L}_i)$
\State \parbox[t]{\linewidth}{$\begin{aligned} \bm{b}_i \gets   \frac{1}{T}\bm{b}_i +\text{vec} \bigl(&-\frac{1}{2}\sum_{j\in\mathcal{N} (i)} \bm{E}_i^{\top} \bm{Z}_j \\&-\frac{\rho}{2} ( \sum_{ j<i,j\in\mathcal{N}(i)}\bm{E}_i^\top\bm{E}_j\bm{L}_j^{k+1}\\&+\sum_{ j>i,j\in\mathcal{N}(i)}\bm{E}_i^\top\bm{E}_j\bm{L}_j^{k})\bigr)\end{aligned}$}
\State $\bm{L}_i^k \gets \mbox{mat}\left (\left ( \frac{1}{T}\bm{A}_i + (\frac{\rho}{2}|\mathcal{N}(i)|+\lambda) \bm{I}_{kd}\right)^{-1}\bm{b}_i\right)$ 
\State  $\bm{Z}^{k+1}_l=\bm{Z}^k_l+\rho(\bm{L}_i^{k+1} - \bm{L}_j^{k+1}) $ (distributed)
\EndWhile
\EndWhile
\EndWhile
\end{algorithmic}
\end{algorithm}

  \section{Theoretical Convergence Guarantees}
 In this section, we provide a proof for convergence of Algorithm~\ref{CoLLA}. W use techniques from \citeauthoryr{Ruvolo2013}, adapted originally from \citeauthoryr{mairal2009online} to demonstrate that Algorithm~\ref{CoLLA} converges to a stationary point of the risk function. In the proof, we assume that the following assumptions hold:

(A) The data distribution has a compact support. This assumption enforces boundedness on $\bm{\alpha}^{(t)}$ and  $\bm{\Gamma}^{(t)}$, and subsequently on $\bm{L}_i$ and $\st{t}$ (see~\cite{mairal2009online} for details).

(B) The LASSO problem in Eq. \eqref{eqn:localbjective} admits a unique solution according to one of uniqueness conditions for LASSO \cite{tibshirani2013lasso}. As a result, the functions $F_i^{(t)}$ are well-defined.

(C) The matrices $\bm{L}_i^\top \bm{\Gamma}^{(t)} \bm{L}_i$ are strictly positive definite. As a result, the functions $F_i^{(t)}$ are all strongly convex.

 Our proof involves two steps. First, we show that the inner loop with variable $k$ in Algorithm~\ref{CoLLA} converges to a consensual solution for all $i$ and all $t$. Next, we prove that the outer loop on $t$ is also convergent, showing that the collectively learned dictionary stabilizes as more tasks are learned. For the first step, we outline the following theorem on the convergence of extended ADMM algorithm.
\begin{theorem}
(Theorem 4.1 of \citeauthoryr{han2012note})
\label{thrm:ADMM convergence}

Suppose we have an optimization problem in the form of Eq. \eqref{eqn:ADMMSparseObjective}, where the functions $g_i(\bm{L}_i)\coloneqq \sum_i F_i^{(t)}(\bm{L}_i)$ are strongly convex with modulus $\eta_i$. Then, for any $0 <\rho < \text{min}_i\{\frac{2\eta_i}{3(N-1)\|\bm{E}_i\|^2}\}$, iterations in Eq. \eqref{eqn:ADMMstepsLa} and Eq. \eqref{eqn:ADMMstepsZ} converge  to a solution of Eq. \eqref{eqn:ADMMSparseObjective}
\end{theorem}

 Note that in Algorithm~\ref{CoLLA}, $F_i^{(t)}(\bm{L}_i)$ is a quadratic function  of $\bm{L}_i$ with a symmetric positive definite Hessian and    thus $g_i(\bm{L}_i)$ as an average of strongly convex functions is also strongly convex. So the required condition for Theorem~\ref{thrm:ADMM convergence} is satisfied and  at each time step, the inner loop on $k$ would converge. We represent the consensual dictionary of the agents after ADMM convergence at time $t=T$ with $\bm{L}_T=\bm{L}_i|_{t=T},\forall i$ (solution obtained via Eq. \eqref{eqn:ADMMstepsZ} and Eq. \eqref{eqn:ADMMSparseObjective} at $t=T$)  and demonstrate that this matrix becomes stable as $t$ grows (outer loop converges), proving overall convergence of the algorithm. More precisely, $\bm{L}_T$ is minimizer of the augmented Lagrangian $ \mathcal{J}_T(\bm{L}_1,\ldots,\bm{L}_N, \bm{Z})$ at $t=T$ and $\bm{L}_1=\ldots=\bm{L}_N$. Also note that  upon convergence of ADMM,  $\sum_i\bm{E}_i\bm{L}_i = \bm{O}$. Hence $\bm{L}_T$ is the minimizer of the following risk function, derived from Eq. \eqref{eqn:AugADMMObjective}:
\begin{equation} \label{eqn:AugADMMObjectiveafterconv}
 \begin{split}
 \hat{\mathcal{R}}_T&(\bm{L})= \frac{1}{T} \sum_{t=1}^T   \sum_{i=1}^N F_i^{(t)}(\bm{L }) +  \lambda\|\bm{L }\|_\textsf{F}^2\enspace.
\end{split}
\end{equation}

We also use the following lemma in our proof \cite{Ruvolo2013}:

\begin{lemma}
\label{lemma: quadratic} The function
$\hat{\mathcal{Q}}_T(\bm{L}) = \hat{\mathcal{R}}_{T}(\bm{L})-\hat{\mathcal{R}}_{T+1}(\bm{L})$  is a Lipschitz function: $\forall$ $\bm{L}$ and $\bm{L}'$, $|\hat{\mathcal{Q}}_{T }(\bm{L}') -\hat{\mathcal{Q}}_{T }(\bm{L})|\le O(\frac{1}{T+1})\|\bm{L}'-\bm{L}\|.$
\end{lemma}
\textbf{Proof}: Note that after algebraic simplifications, we can conclude: $\hat{\mathcal{Q}}_T(\bm{L})=\bigl( \frac{1}{T(T+1)} \sum_{t=1}^T   \sum_{i=1}^N F_i^{(t)}(\bm{L })\bigr)-\frac{1}{T+1}F_i^{(T+1)}$. Now note that the functions $F_i^{(t)}(\bm{L }) $  are quadratic forms with positive  definite Hessian matrices and hence are Lipschitz functions, all with Lipschitz parameters  upper-bounded by the largest eigenvalue of  all Hessian matrices. Using the definition for a Lipschitz function, it is  easy to demonstrate that $\hat{\mathcal{R}}_T(\cdot)$  is also Lipschitz with Lipschitz parameter $O(\frac{1}{T+1})$, because of averaged quadratic form terms in Eq. \eqref{eqn:AugADMMObjectiveafterconv}.

Now we can prove the convergence of Algorithm~\ref{CoLLA}:
\begin{lemma}
\label{lemm2:convergence}
$\bm{L}_{T+1}-\bm{L}_T=O(\frac{1}{T+1})$,   showing that Algorithm 1 converges to a stable dictionary as $T$ grows large. 
\end{lemma}

\textbf{Proof:} First note that $\hat{\mathcal{R}}_T(\cdot)$ is a strongly convex function for all $T$. Let $\eta_T$ be the strong convexity modulus. From the definition, for two points $\bm{L}_{T+1}$ and $\bm{L}_T$, we have: $\hat{\mathcal{R}}_{T }(\bm{L}_{T+1})\ge \hat{\mathcal{R}}_{T}(\bm{L}_{T })+\nabla\hat{\mathcal{R}}_{T}^\top(\bm{L}_{T})(\bm{L}_{T}-\bm{L}_{T+1})+\frac{\eta_T}{2}\|\bm{L}_{T+1}-\bm{L}_T\|_\textsf{F}^2$. Since $\bm{L}_{T }$ is minimizer of $\hat{\mathcal{R}}_T(\cdot)$:
\begin{equation} \label{eqn:proof1}
 \begin{split}
\hat{\mathcal{R}}_{T }(\bm{L}_{T+1}) -\hat{\mathcal{R}}_{T }(\bm{L}_{T })\ge\frac{\eta_T}{2}\|\bm{L}_{T+1}-\bm{L}_T\|_\textsf{F}^2\enspace.
\end{split}
\end{equation}
On the other hand, from Lemma~\ref{lemma: quadratic}:
\begin{equation} \label{eqn:proof3}
 \begin{split}
&\hat{\mathcal{R}}_{T }(\bm{L}_{T+1}) -\hat{\mathcal{R}}_{T }(\bm{L}_{T })=\hat{\mathcal{R}}_T(\bm{L}_{T+1})-\hat{\mathcal{R}}_{T+1}(\bm{L}_{T+1})+\\&
\hat{\mathcal{R}}_{T+1}(\bm{L}_{T+1})-\hat{\mathcal{R}}_{T+1}(\bm{L}_T)+ 
\hat{\mathcal{R}}_{T+1}(\bm{L}_T)-\hat{\mathcal{R}}_T(\bm{L}_T) \le  \\&\hat{\mathcal{Q}}_T(\bm{L}_{T+1})-\hat{\mathcal{Q}}_T(\bm{L}_T)\le O(\tfrac{1}{T+1})\|\bm{L}_{T+1}-\bm{L}_T\|\enspace.
\end{split}
\end{equation} 
Note that the first two terms in the second line in the above as a whole is negative since $\bm{L}_{T+1}$ is the minimizer of $\hat{\mathcal{R}}_{T+1}$.
Now combining \eqref{eqn:proof1} and \eqref{eqn:proof3}, it is easy to show  that $\|\bm{L}_{T+1}-\bm{L}_T\|_\textsf{F}^2\le O(\tfrac{1}{T+1})$, concluding the proof:%\hfill$\blacksquare$
  \begin{equation} \label{eqn:proof2}
  \begin{split}
  \|\bm{L}_{T+1}-\bm{L}_T\|_\textsf{F}^2\le O(\tfrac{1}{T+1})  \hspace{4mm}\blacksquare
 \end{split}
 \end{equation} 
 
Thus, Algorithm~\ref{CoLLA} converges as $t$ increases. We also show that the distance between $\bm{L}_T$ and the set of stationary points of the agents' true expected costs  $\mathcal{R}_T=\mathbb{E}_{\Xt{t}\sim\mathcal{D}^{(t)}}(\hat{\mathcal{R}}_T)$ converges almost surely (a.s.) to $0$ as $T\rightarrow \infty$. We use two theorems from \citeauthoryr{mairal2009online}
for this purpose.
\begin{theorem}
\cite{mairal2009online}
\label{thrm:ELLAconv2}
Consider the empirical risk function $\hat{q}_T(\bm{L}) =\frac{1}{T} \sum_{t=1}^TF^{(t)}(\bm{L}) + \lambda\|\bm{L}\|_\textsf{F}^2$ with $F^{(t)}$ as defined in \eqref{eqn:localbjective} and the true risk function \mbox{$q_T(\bm{L})=\mathbb{E}_{\Xt{t}\sim\mathcal{D}^{(t)}}(\hat{g}(\bm{L}))$}, and make assumptions (A--C). Then both risk functions converge a.s. as $\lim_{T\rightarrow \infty} \hat{q}_T(\bm{L})-q_T(\bm{L})=0$.
\end{theorem}
Note that we can apply this theorem on $\mathcal{R}_T$ and $ \hat{\mathcal{R}}_T$, because the inner sum in  \eqref{eqn:AugADMMObjectiveafterconv} does not violate the  assumptions of Theorem~\ref{thrm:ELLAconv2}. This is because the functions $g_i(\cdot)$ are all well-defined and are strongly convex with strictly positive definite Hessians (the  sum of positive definite matrices is positive definite). Thus,  $\lim_{T\rightarrow \infty} \hat{\mathcal{R}}_T-\mathcal{R}_T=0$ a.s.
 \begin{theorem}
\cite{mairal2009online}
\label{thrm:ELLAconv3}
Under assumptions (A--C), the distance between the minimizer of $\hat{q}_T(\bm{L})$ and the stationary points of $q_T(\bm{L})$ converges almost surely to zero.
\end{theorem}
Again, this theorem is applicable on $\mathcal{R}_T$ and $ \hat{\mathcal{R}}_T$ and thus Algorithm~\ref{CoLLA} converges to a stationary point of the true risk.

 \textbf{Computational Complexity}~~
At each time-step, each agent computes the optimal ridge parameter  $\bm{\alpha}^{(t)}$ and the Hessian matrix $\bm{\Gamma}^{(t)}$ for the received task. This has a cost of $O(\xi(d,M))$, where $\xi()$ depends on the base   learner.  The cost of updating $\bm{L}_i$ and $\st{t}_i$ alone is $O(u^2d^3)$ \cite{Ruvolo2013}, and so the cost of updating all local dictionaries by the agents  is $O(Nu^2d^3)$.  Note that this step is performed $K$ times in each time-step. Finally, updating the dual variable requires a cost of $eud$. This leads to overall cost of  $O(N\xi(d,M)+ K(Nu^2d^3+eud)  )$, which is independent of $T$ but accumulates as more tasks are learned. We can think of the factor $K$ in the second term as communication cost because in a centralized scheme we would not need these repetitions which requires sharing the local bases with the neighbors. Also, note that if the number of data points per task is big enough, it certainly is more costly to send data to a single server and learn the tasks in a centralized optimization scheme.  

 \begin{figure*}[t!]
    \centering
\begin{subfigure}[b]{0.25\textwidth}
        \includegraphics[width=2 in,height=1.45in]{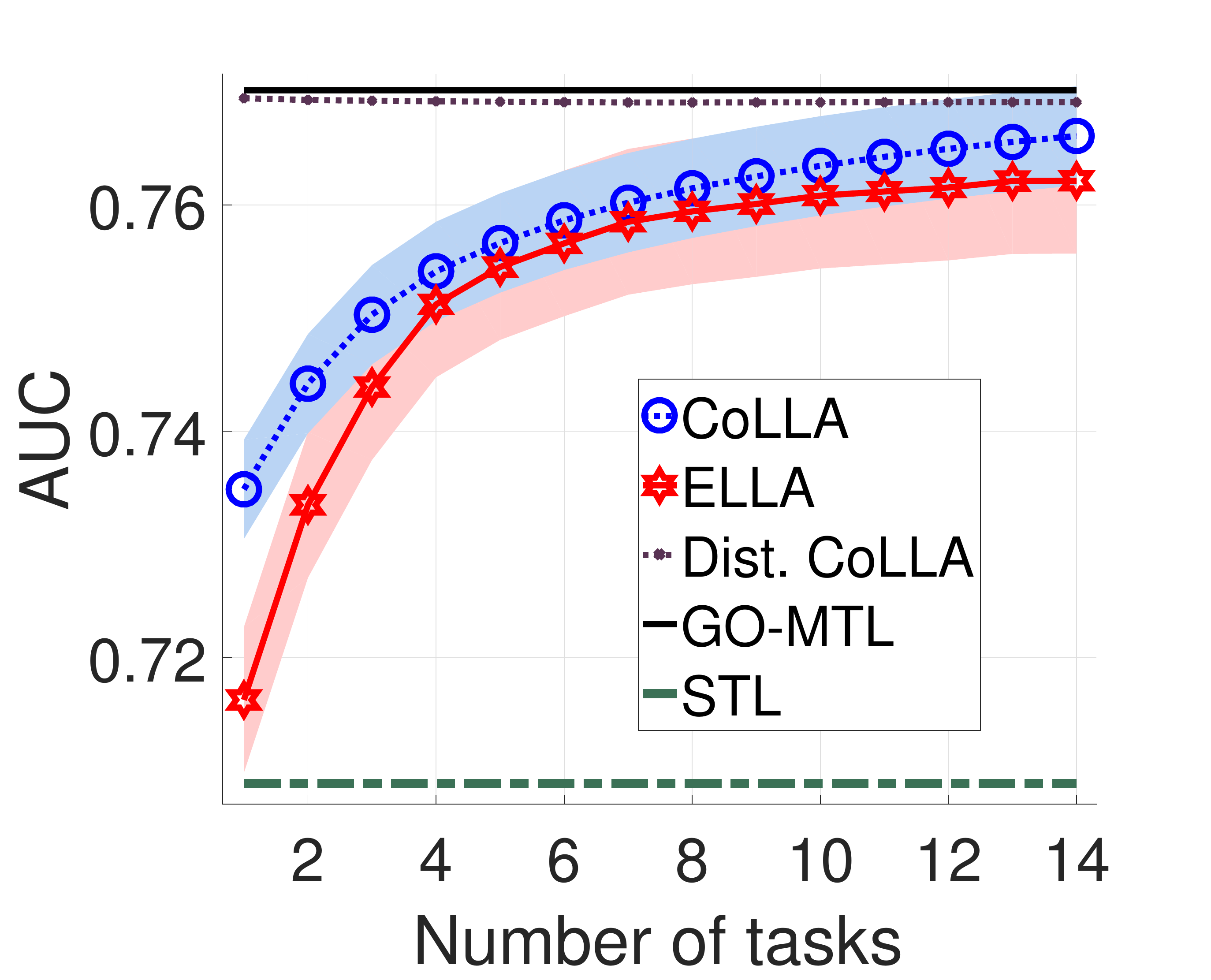}
        \caption{Land Mine}
        \label{fig:Landmine}
    \end{subfigure}%
    \begin{subfigure}[b]{0.25\textwidth}
        \includegraphics[width=2 in,height=1.45in]{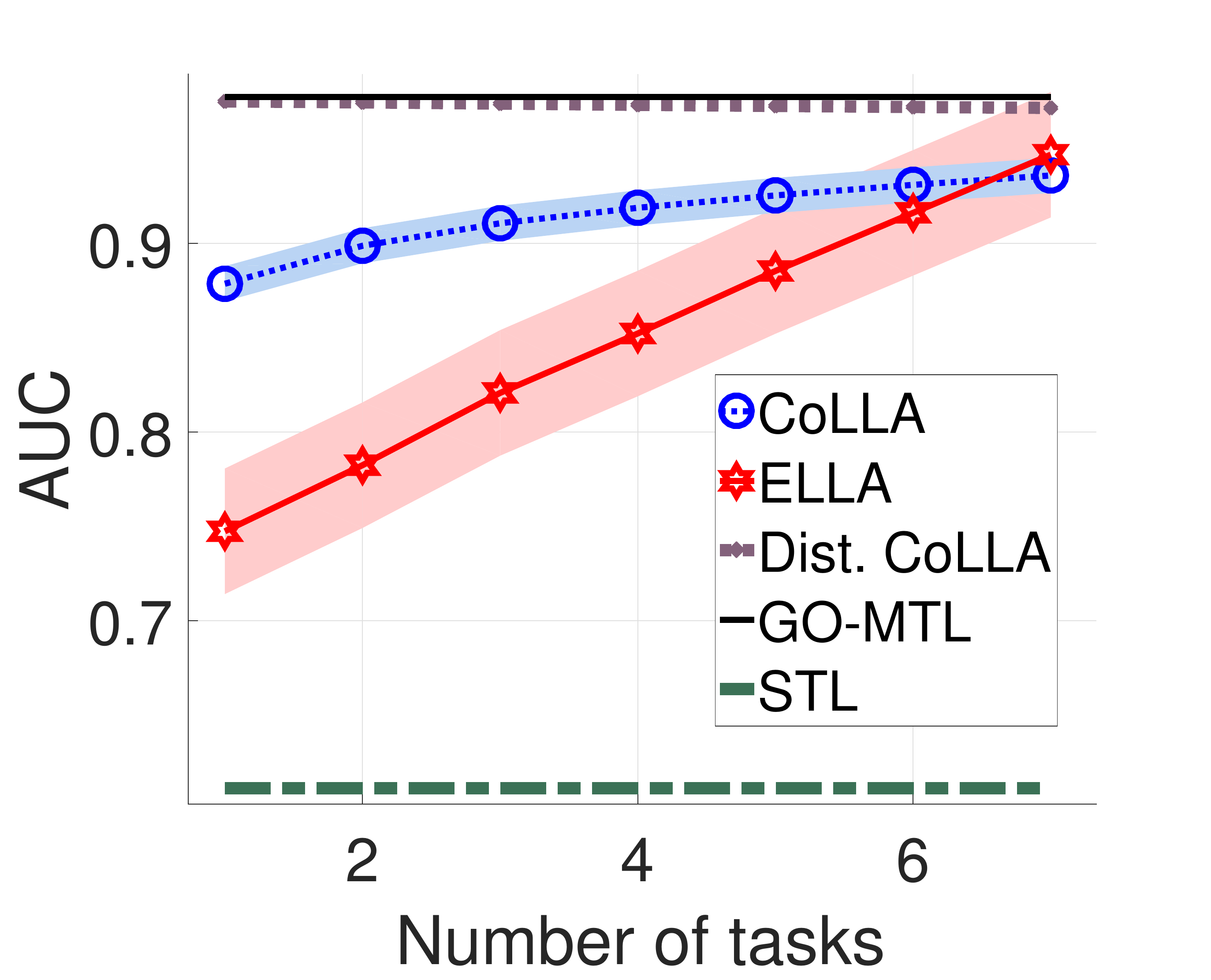}
        \caption{Facial Expression}
        \label{fig:School}
            \end{subfigure}%
    \begin{subfigure}[b]{0.25\textwidth}
       \includegraphics[width=2 in,height=1.45in]{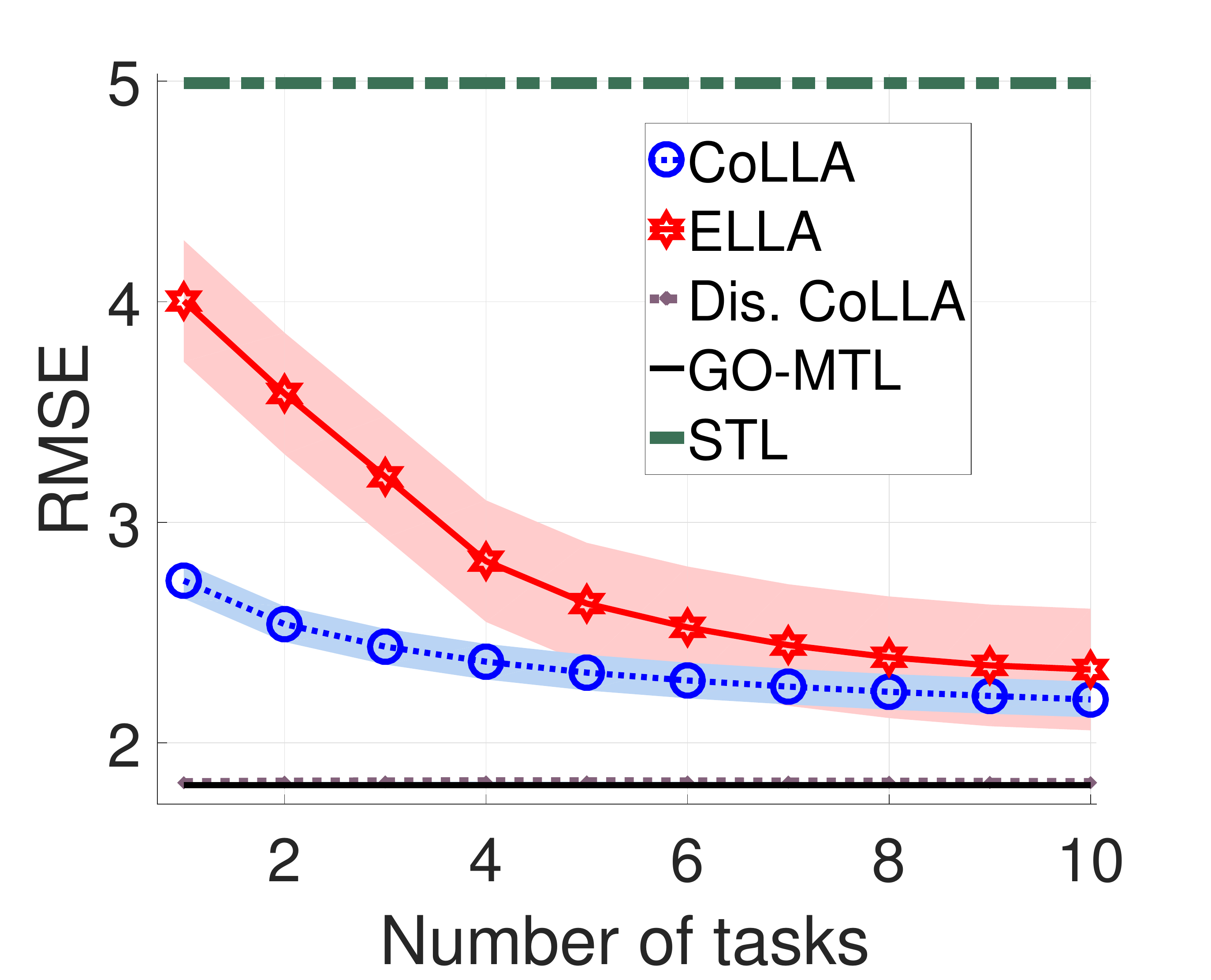}
        \caption{Computer Survey}
        \label{fig:Computer}
            \end{subfigure}%
    \begin{subfigure}[b]{0.25\textwidth}
       \includegraphics[width=2 in,height=1.45in]{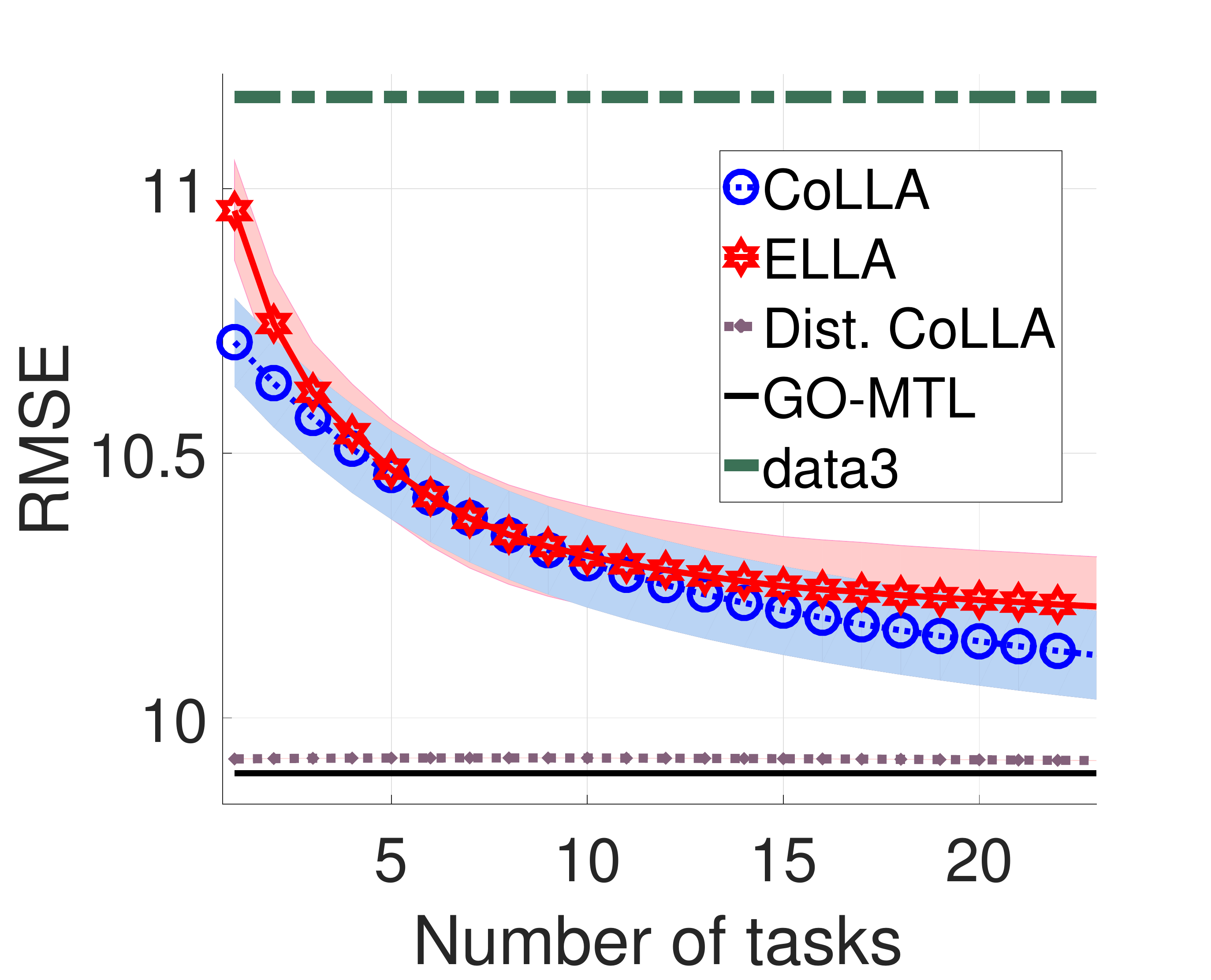}
        \caption{London Schools}
        \label{fig:Facial}
    \end{subfigure}
    \vspace{-1.8em}
    \caption{Performance of distributed (dotted lines), centralized (solid), and single-task learning
(dashed) algorithms on benchmark datasets. Shaded region shows standard error (Best viewed in color.)  }
    \label{fig:results1}
\end{figure*}

\begin{figure*}[t!]
    \centering
\begin{subfigure}[b]{0.25\textwidth}
        \includegraphics[width=1.45 in,height=1.45in] {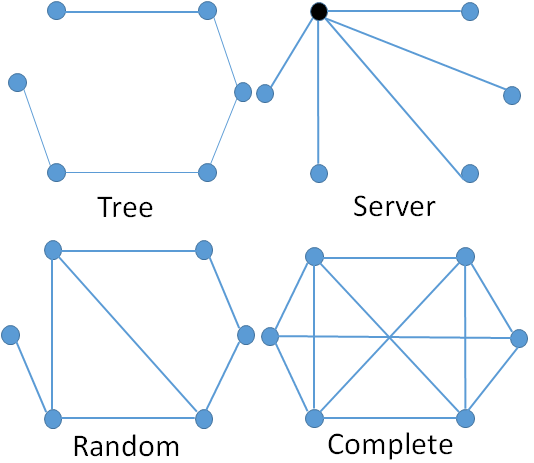}
        \caption{Graph Structures}
        \label{fig:LMgraphs}
    \end{subfigure}%
    \begin{subfigure}[b]{0.25\textwidth}
        \includegraphics[width=2 in,height=1.55in]{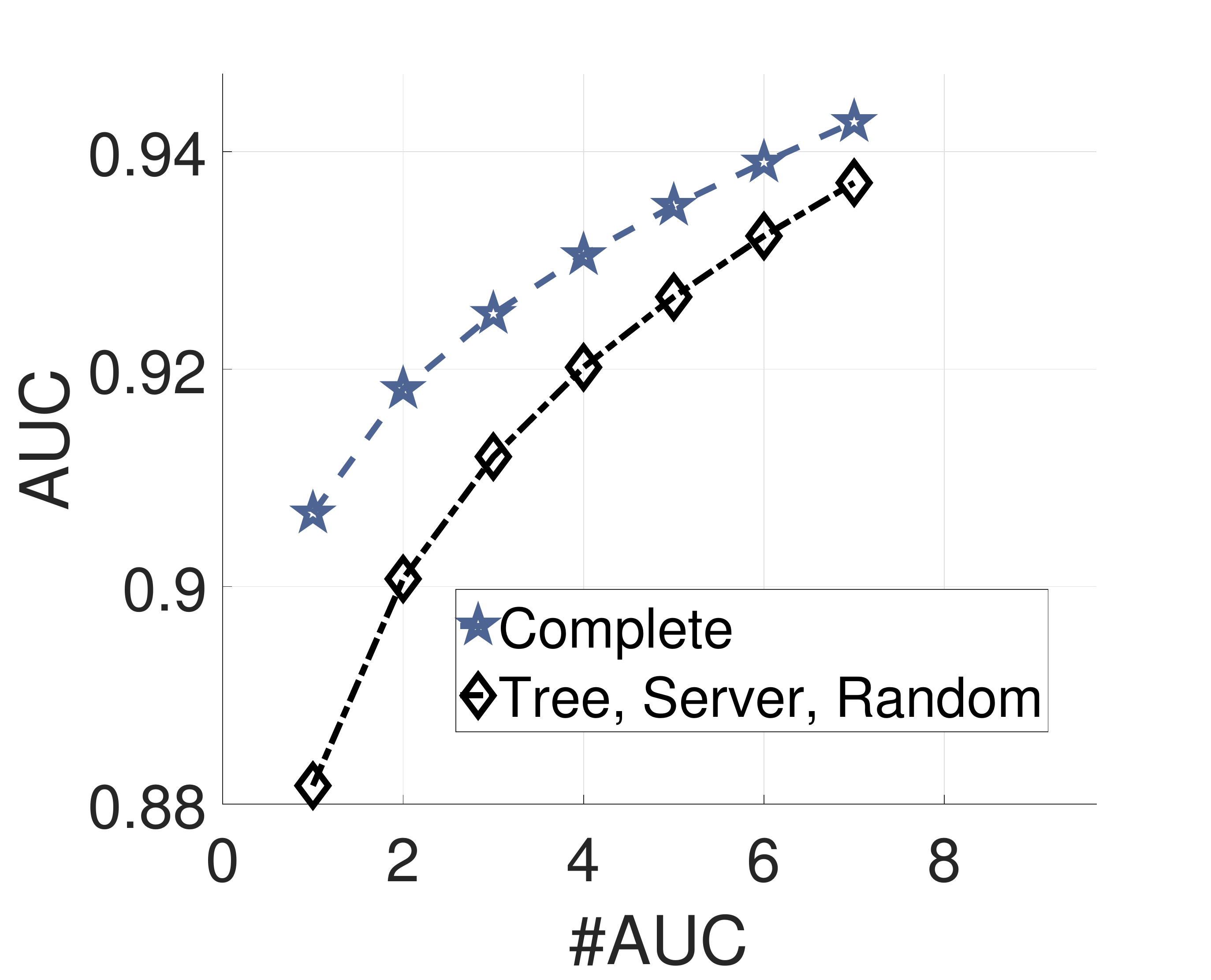} 
        \caption{Facial Expression}
        \label{fig:Schoolgraph}
            \end{subfigure}%
    \begin{subfigure}[b]{0.25\textwidth}
       \includegraphics[width=2 in,height=1.55in] {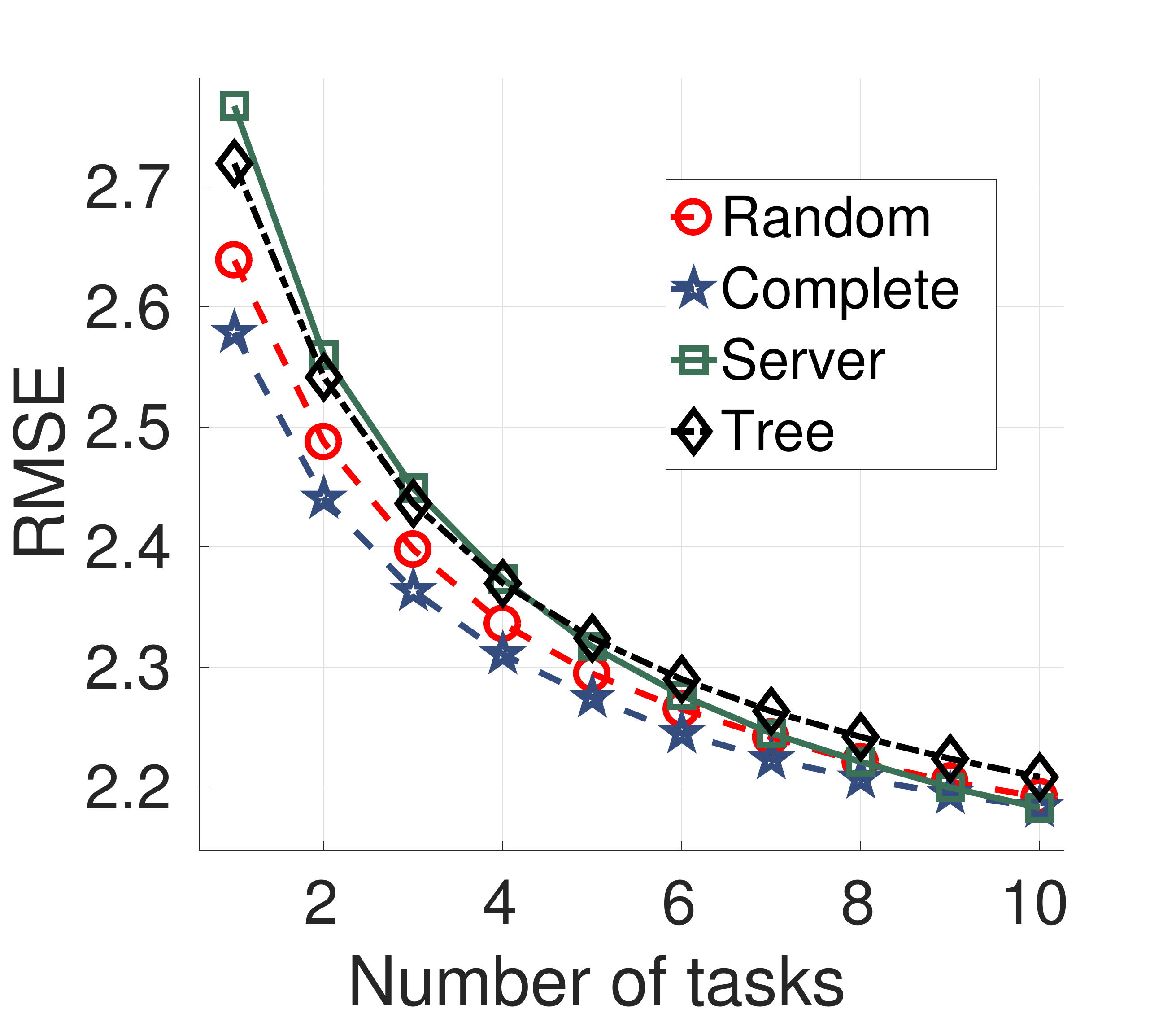}
        \caption{Computer Survey}
        \label{fig:Computergraphs}  
            \end{subfigure}%
    \begin{subfigure}[b]{0.25\textwidth}
       \includegraphics[width=2 in,height=1.55in]{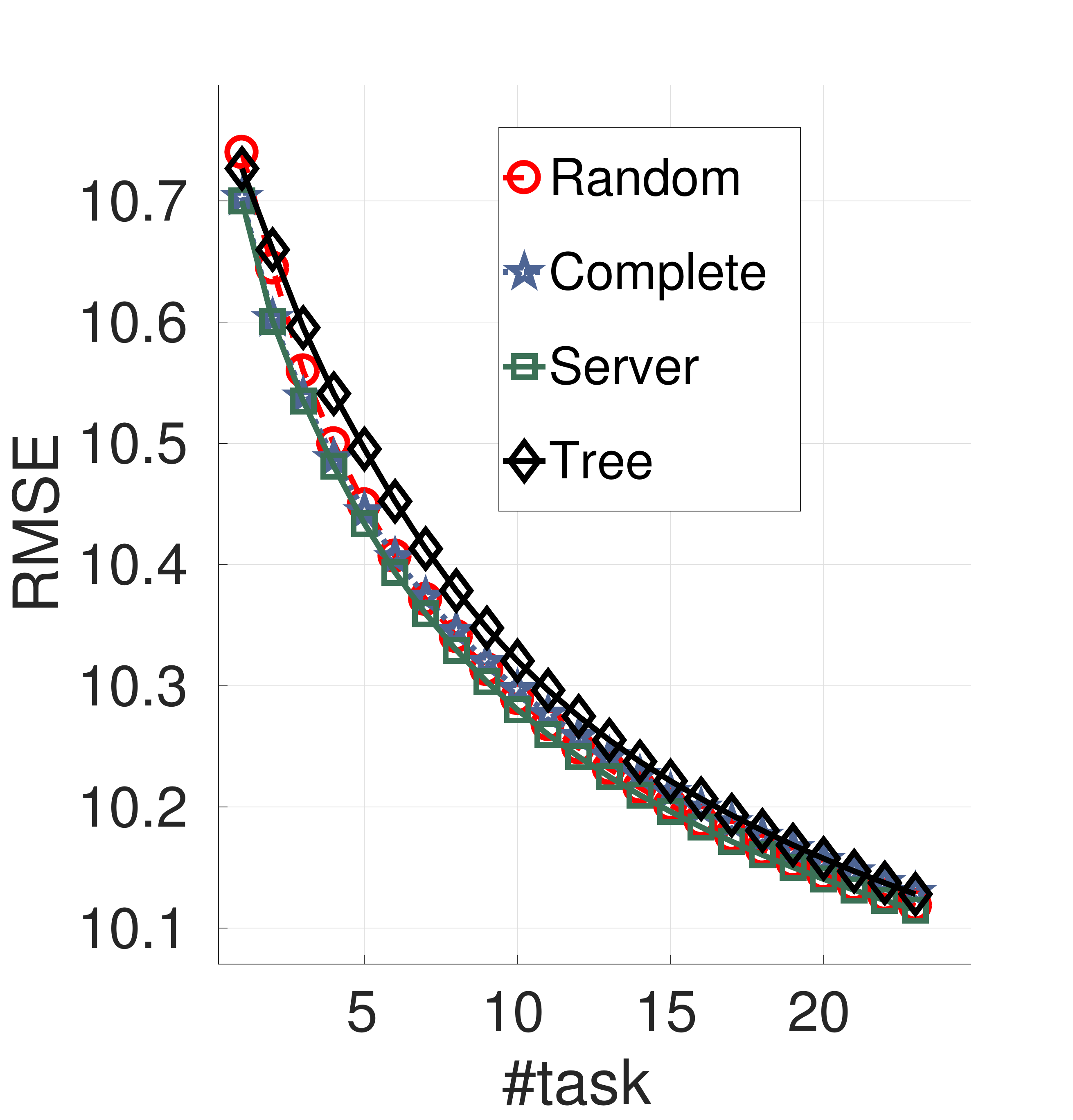} 
        \caption{London Schools}
        \label{fig:Facialgraphs}
    \end{subfigure}
    \vspace{-1.8em}
    \caption{Performance of CoLLA given various  graph structures (a) for three data sets (b--d).  }
    \label{fig:results2}
\end{figure*}

 \section{Experimental Results}
We validate our algorithm by comparing it against: 1) STL, a lower-bound to measure the effect of positive transfer from the learned tasks, 2)   ELLA, to demonstrate  that collaboration between the agents improves overall performance compared to ELLA, 3) offline CoLLA, as an upper-bound to our online distributed algorithm, and finally 4) GO-MTL, as an absolute upper-bound  (since GO-MTL is a batch MTL method) to assess the performance of our algorithm from different perspectives. Throughout all experiments, we present and compare the average performance of all agents.

\subsection{Datasets}
We used four benchmark MTL datasets in our experiments, including two classification  and two regression datasets: 1)  land mine detection in radar images~\cite{Carin2007}, 2) facial expression identification   from photographs of a subject's face ~\cite{valstar2011first}, 3) predicting London students' scores using school-specific and student-specific features~\cite{Argyriou}, and 4) predicting   ratings of customers for different computer models \cite{lenk1996hierarchicaln}. Below we describe each dataset. Note that for each dataset, we assume that the tasks are distributed equally among the agents. We used different numbers of agents across the datasets to explore various network sizes of the multi-agent system in our framework.
 
\textbf{Land Mine Detection:} This dataset consists of binary classification tasks to detect  whether a geographical region contains a land mine from a radar image. There are 29 tasks, each corresponding to a different geographical region, with a total 14,820 data points. Each data point consists of nine  features, including four moment-based features, three correlation-based, one energy-ratio, and one spatial variance feature (see Xue et. al.~\cite{Carin2007} for details), all automatically extracted from radar images. We  added a bias term as a 10$^{th}$ feature. The dataset has a natural dichotomy between foliated and  dessert regions. We assumed there are two collaborating agents, each dealing solely with one geographical region type.

\textbf{Facial Expression Recognition:} This dataset consists of binary facial expression recognition tasks \cite{valstar2011first}. We followed \citeauthoryr{Ruvolo2013} and chose tasks detecting three facial action units   (upper lid raiser, upper lip raiser, and lip corner pull) for seven  different subjects, resulting in 21 total tasks, each with 450--999 data points. A Gabor pyramid scheme is used to extract a total number of 2,880 Gabor features from images of a subject's face (see~\citeauthoryr{Ruvolo2013}  for details). Each data point  consists of the first 100 PCA components of  these Gabor features. We used three  agents, each of which learns seven randomly selected tasks. Given that facial expression recognition is a core task for personal assistant robots, each agent can be considered a   personal service robot that interacts with few people in a specific environment.

\textbf{London Schools:}  This dataset was provided by the Inner London Education Authority. It consists of examination scores  of 15,362 students (each assumed to be a data point) in 139 secondary schools  (each assumed to be a single task) during three academic years. The goal is to predict the score of students of each school  using provided features as a regression problem. We used the same 27 categorical features as described by \citeauthoryr{kumar2012learning}, consisting of eight school-specific features and 19  student-specific features, all encoded as binary features. We also added a feature to account for the bias term. For this dataset, we considered six agents and allocated 23 tasks randomly to each agent.

\textbf{Computer Survey:}
The goal in this dataset is to predict the likelihood of purchasing one of 20 different computers by 190 subjects. Each subject is assumed to be a task and its ratings determines the task data points. Each data point consists of 13 binary features, e.g. guarantee, telephone hot line, etc. (see \citeauthor{lenk1996hierarchicaln}~\citeyear{lenk1996hierarchicaln} for details). We  added a feature to account for the  bias term. The output is a rating on a scale 0--10 collected in a survey from the subjects. We considered 19 agents and randomly allocated ten tasks to each.

\subsection{Evaluation methodology}
For each experiment, we randomly split the  data for each task evenly into  training and testing sets. We performed 100 learning trials on training sets and reported the average performance on the testing sets for these trials as well as the performance variance. For the online settings (CoLLA and ELLA), we also  randomized the order of task presentation in each trial to rule out biases in the order of learning  tasks. For the offline settings (GO-MTL, Dist. CoLLA, STL), we reported the average asymptotic performance on all task because all tasks are presented and learned simultaneously. We used brute force search to cross validate the  parameters $u$, $\lambda$, $\mu$, and $\rho$ for each dataset; these parameters were selected to maximize the performance on a validation set for each algorithm independently. Parameters $\lambda$, $\mu$, and $\rho$ are selected from the set $\{10^{n}|-6\le n\le 6\}$ and $u$ from $\{1,\ldots,\text{max}(10,\frac{T}{4})\}$ (Note that $u \ll T$).

 \textbf{Quality of agreement among the agents:} The inner loop in Algorithm~\ref{CoLLA} implements information exchange between the agents. For effective collective learning, agents need to come to an agreement at each time-step which is guaranteed by ADMM if $K$ is chosen large enough. During our experiments, we noticed that   initially $K$ needs to be fairly large but as more tasks are learned, it can be decreased over time  $K\propto K_1+K_2/t$ without considerable change in performance ($K_1\in \mathbb{N}$ is generally small and $K_2\in \mathbb{N}$ is large). This is expected because the learned tasks by all agents are homogeneous and hence as more tasks are learned, knowledge transfer from previously learned tasks makes local dictionaries closer.

 For the two regression problems, we used standard root mean-squared error (RMSE) on  the  testing set to measure performance of the algorithms. For the two classification problems, we used the area under the ROC curve (AUC) to measure performance. We used AUC because both classification datasets have highly unbiased distributions, making RMSE less informative. Unlike AUC, RMSE is agnostic to the trade-off between false-positives and false-negatives, which can vary in terms of importance in different applications.

 \subsection{Results}   

We clarify that  In the result section, the number of learned tasks is equal for both COLLA and ELLA. While the x-axis is the number of tasks learned by a single agent, but we are reporting the average performance by all agents. We used this visualization, because the benefit of collaborative scheme can be seen on average for all agents.

For the first   experiment  on CoLLA, we assumed a minimal linearly connected tree which allows for information flow among the agents~\mbox{$\mathcal{E}=\{(i,i+1)|1\le i\le N\}$}.  Figure~\ref{fig:results1} compares CoLLA against ELLA  (which does not use collective learning), GO-MTL, and single-task learning. Note that at each time  step  $t$, we report  the average performance of online algorithms on all learned tasks up to that instance in the vertical axis. Thus the horizontal axis denotes the number of learned tasks by each agent solely for online learning setting. ELLA can be considered as a special case of COLLA with an edgeless graph topology (no communication). This would allow us to assess whether consistently positive/ transfer has occurred. A progressive   increase in the average performance on the learned tasks demonstrates that positive transfer has occurred and allows plotting  learning curves.  
 Moreover, we also performed an offline distributed batch MTL optimization of Eq.~\eqref{eqn:ADMMSparseObjective}, i.e. offline CoLLA. For comparison, we plot the learning curves for the online settings and the average asymptotic performance on all tasks for offline settings in the same plot. The shaded regions on the plots denote the standard error for 100 trials.  

 \begin{table}[t!]
 \centering
\begin{tabular}{c|cccc }   
\centering
\diaghead{MethodDataset}{Method}{Dataset}           & LM & LS & CS & FE   \\
\hline
 CoLLA			& 6.87 &  29.62 & 51.44   & 40.87\\
 \hline
  ELLA        &      6.21 &     29.30 & 37.99  & 38.69 \\
\hline
Dist. CoLLA           & 32.21 & 37.3 & 61.71   & 59.89 \\
\hline
GO-MTL		  & 8.63 & 32.40 &  61.81   & 60.17\\
\hline
\end{tabular}
\caption{ Jumpstart comparison (improvement in percentage) on the Land Mine (LM), London Schools (LS), Computer Survey(CS), and Facial Expression (FE) datasets}
\label{tab:table1}
\end{table}

 Figure~\ref{fig:results1} shows  that collaboration among  agents  expectedly improves lifelong learning, both in terms of learning speed and asymptotic performance, to a level that is not feasible for a single lifelong learning agent.  Also,  performance of offline CoLLA is comparable with GO-MTL, demonstrating that our algorithm can be used   as a distributed MTL algorithm. As expected, both CoLLA and ELLA lead to the same asymptotic performance because they solve the same optimization problem. These results demonstrate the effectiveness of our algorithm for both offline and online optimization settings. We also measured the improvement in the initial performance on a new task due to transfer (the ``jumpstart'' \cite{taylor2009transfer}) in Table~\ref{tab:table1}, highlighting COLLA's effectiveness in collaboratively learning knowledge bases suitable for transfer. This result accords with intuition because collaboration is most effective in learning initial tasks.

We conducted  a second set of experiments to study the effect of  the  communication mode (i.e., the graph structure) on distributed lifelong learning. We performed experiments on   four graph structures visualized in Figure~\ref{fig:LMgraphs}: tree, server (star graph), complete, and random. The server graph structure connects all agents (slave agents) through a central server (master agent) (depicted in black in the figure), and the random graph was formed by randomly selected half of the edges of a complete graph while still ensuring that the resulting graph was connected.   Note that some of  these structures coincide when the network is small (for this reason, results on the land mine dataset are not presented for this second experiment). Performance results for these structures on the London schools, computer survey, and facial expression recognition datasets are presented in Figures~\ref{fig:Schoolgraph}--\ref{fig:Facialgraphs}.  Note that for facial recognition data set, results for the only two possible structures are presented. From these figures, we can roughly conclude that for network structures with more edges, the learning rate is faster. Intuitively, this empirical result signals that more communication and collaboration between the agents can increase learning speed.

 \section{Conclusion} \label{sect:Conclusion}
   We proposed a distributed optimization algorithm for enabling collective multi-agent lifelong learning. Collaboration among the agents not only improves the asymptotic performance on the learned tasks, but enables the agent to learn faster (i.e. using less data to reach a specific performance threshold). Our  experiments demonstrated that  the proposed algorithm outperforms other alternatives on a variety of MTL regression and classification problems. Extending the proposed framework to a network of asynchronous agents with dynamic links is a potential future direction to improve the applicability of the algorithm on real world problems.
 
 %\section*{Acknowledgments}

%Omitted for blind review.
 
%%%%%%%%%%%%%%%%%%%%%%%%%%%%%%%%%%%%%%%%%%%%%%%%%%%%%%%%%%%%%%%%%%%%%%%%%%%%%%%%%%%%%%%%%%%%%%%%%%%%%%%%%
%% bibliography: see CFP for number of per mitted pages

\bibliographystyle{aamas18}  % do not change this line!
\bibliography{aamas18}  % put name of your .bib file here

\end{document}